\documentclass[letterpaper, 10 pt, conference]{ieeeconf}  
\usepackage{graphicx}
\usepackage{amsmath}
\usepackage{amssymb}
\usepackage{placeins}
\usepackage{booktabs}       
\usepackage{tabularx}       
\usepackage{threeparttable} 
\usepackage{multirow}
\usepackage{float}
\usepackage{xurl}
\usepackage{balance}
\usepackage{censor}
\RestartCensoring

\IEEEoverridecommandlockouts                              

\title{\LARGE \bf
Context-Aware Force Estimation for Deformable Tool Manipulation in Robotic Environmental Swabbing via Few-Shot Continual Adaptation
}

\author{Siavash Mahmoudi$^{1}$, Chaitainya Kuppar Reddy$^{2}$, Yang Tian$^{1}$, and Dongyi Wang$^{1,2}$
\thanks{*This work is supported by the project award no. 2023-70442-39232, from the
U.S. Department of Agriculture's National Institute of Food and Agriculture.}
\thanks{$^{1}$Authors are with the Department of Biological and Agricultural Engineering, University of Arkansas, Fayetteville, AR 72701, USA. Email: siavashm@uark.edu}
\thanks{$^{2}$Author is with the Department of Food Science, University of Arkansas, Fayetteville, AR 72701, USA. Email: dongyiw@uark.edu}
}

\begin{document}

\maketitle
\thispagestyle{empty}
\pagestyle{empty}

\begin{abstract}
Robotic surface swabbing requires sustained interaction between a compliant tool and heterogeneous environments, where accurate estimation of tip-level contact force is critical for consistent sampling performance. However, deformable tool dynamics introduce nonlinear viscoelastic hysteresis that decouples wrist-mounted force measurements from true contact forces, while tool-integrated sensors are impractical for deployment due to sterility and disposability constraints.
This paper presents a data-driven framework for contact force estimation in Deformable Tool Manipulation (DTM) that leverages proprioceptive sensing without requiring explicit physical models or permanent embedded sensing hardware at the tool tip. A recurrent architecture is first identified through a comparative evaluation of temporal models, where a compact LSTM achieves the lowest estimation error and sub-millisecond inference latency. To address generalization across unseen surfaces and tool compliance conditions, we introduce a parameter-isolated few-shot adaptation strategy that augments a frozen recurrent backbone with low-dimensional context embeddings using feature-wise linear modulation (FiLM).
Experiments on a UR5e platform across nine tool–surface interaction regimes demonstrate that the proposed approach significantly improves robustness under domain shift, reducing zero-shot estimation error by up to 63\% while preserving baseline performance without catastrophic forgetting. These results show that separating shared deformation-history dynamics from domain-specific conditioning enables reliable force estimation for DTM in non-stationary environments.
\end{abstract}

\section{INTRODUCTION}

Deformable interaction is a long-standing challenge in robotics and has been extensively studied under the umbrella of deformable object manipulation (DOM), with applications such as cloth folding, rope handling, and food manipulation \cite{zhu2022challenges}. In these problems, a rigid robot end-effector interacts with a deformable object, and the primary challenge lies in modeling and controlling the object deformation \cite{yin2021modeling}.

\begin{figure}[t]
    \centering
    \includegraphics[width=1\linewidth]{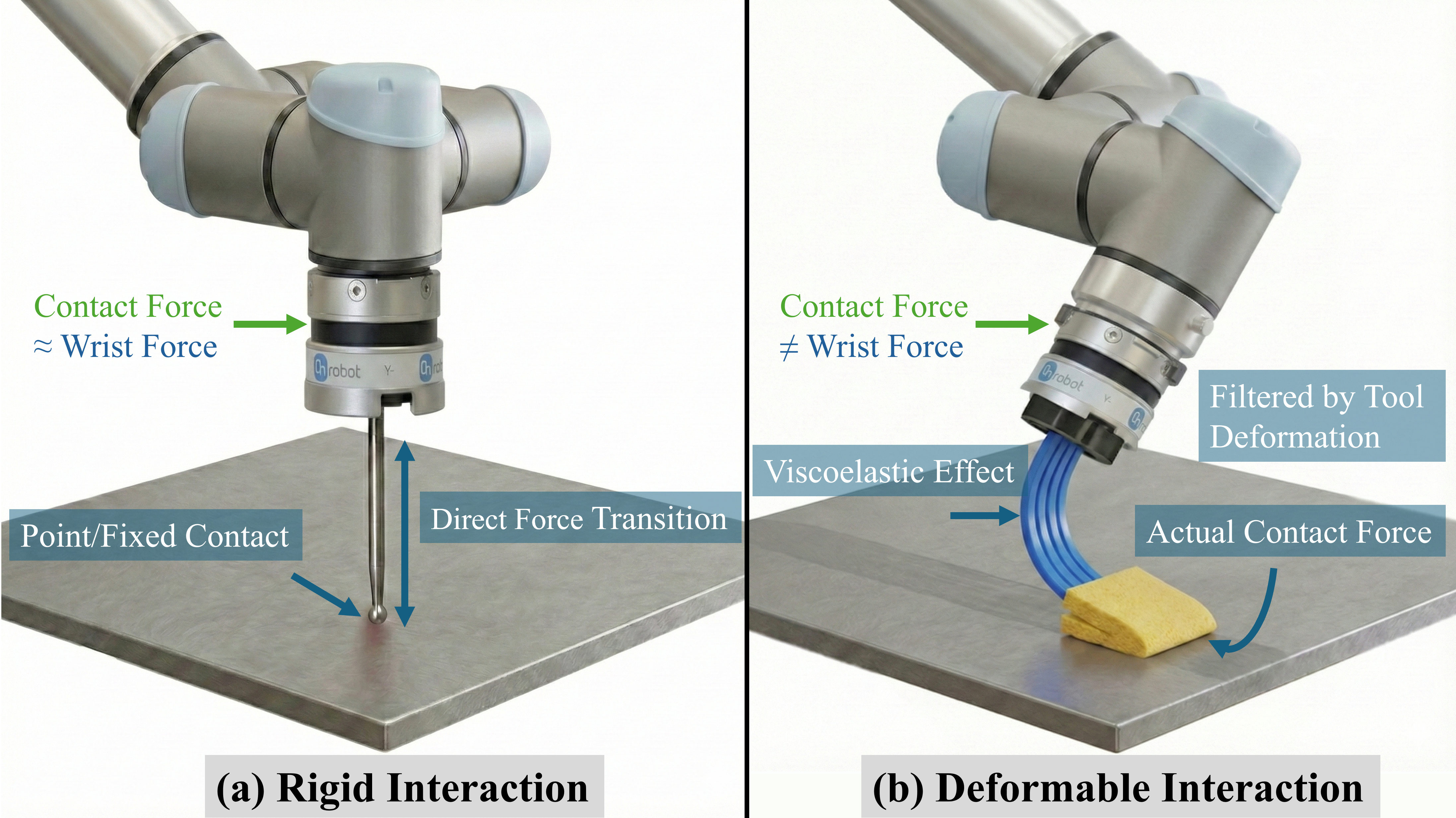}
    \caption{Force transmission dynamics. (a) Rigid tools show direct correspondence between wrist and contact forces. (b) Tool deformation decouples wrist sensing from tip forces.}
    \label{fig:rvsf}
    \vspace{-20pt}
\end{figure}

Manipulation tasks in which a robot operates through a deformable tool constitute a distinct and often more challenging class of problems, as the robot must interact with the environment indirectly through a compliant, deformable medium rather than directly manipulating a deformable object \cite{sloth2022towards}. As illustrated in Fig. \ref{fig:rvsf}, in such settings, the tool itself undergoes continuous internal deformation during contact, causing the interaction forces at the tool–environment interface to be filtered, delayed, and decoupled from the forces measured at the robot wrist. This sensing and control mismatch makes accurate force estimation particularly difficult \cite{bazaei2012fundamental}.
DTM is central to a wide range of practical applications, including wiping, brushing, polishing, surface inspection, surgical swabbing, and environmental sampling, where sustained contact and precise force regulation are essential for task effectiveness and safety. Unlike rigid tool manipulation, where interaction forces can often be approximated by point or planar contacts governed by relatively simple friction models \cite{fernandez2024stark}, force generation in deformable tool interactions such as viscoelastic polishing pads, porous sponge swabs, or bristle-based brushes is strongly influenced by material properties, deformation history, and interaction velocity \cite{mahmoudi2025data}, introducing complex, high-dimensional interaction dynamics that are difficult to observe and model explicitly \cite{unger2024prosip}. The effective state of the tool is no longer defined solely by Cartesian position and orientation, but by its internal deformation, stress distribution, and, in some cases, fluid absorption or dissipation within the material. This internal deformation decouples the forces measured at the robot wrist from the forces actually applied at the tool–surface interface, significantly limiting the reliability of conventional wrist-mounted force–torque sensing for deformable contact tasks.

Deformable tool interactions exhibit three coupled properties that challenge classical modeling: geometric nonlinearity, where contact area and pressure distribution evolve as the tool compresses, splays, or reorients \cite{mahmoudi2025evaluation}; hysteretic behavior, where internal energy dissipation causes forces during loading to differ from those during unloading at identical configurations \cite{kowalewski2025sccrub}; and rate-dependent stiffness, where apparent material properties depend on interaction velocity due to limited internal relaxation during rapid motions \cite{kim2010parametric}. Together, these factors render the tool's effective state high-dimensional, defined not only by pose but also by internal stress distribution and deformation history, precluding closed-form force-displacement models and traditional physics-based approaches. The challenges also make accurate estimation and regulation of contact force critical \cite{shin2025optimal}. 

Although learning-based approaches offer a compelling alternative by capturing complex interaction dynamics directly from data \cite{tsuji2025survey}, they face significant challenges in terms of data efficiency and generalization. Recent work demonstrates that neural models can approximate force-motion mappings from raw sensor inputs without explicit physical modeling \cite{shan2023fine}; however, these methods typically require extensive data collection, often exceeding 1000 trials per surface configuration, and exhibit brittleness when faced with distribution shifts such as variations in surface material or tool compliance \cite{tobin2017domain}. Furthermore, adapting a pre-trained model to these new interaction regimes remains difficult: standard fine-tuning not only requires extensive new data, but also risks catastrophic forgetting, where the model’s proficiency on previously learned tasks degrades \cite{dong2024mitigating}.

This paper presents a data-driven framework for contact force estimation in DTM, demonstrated on robotic environmental surface swabbing.  Effective swabbing requires sustained sliding contact between a compliant swabbing tool and heterogeneous surfaces, where the contact force must be precisely regulated to ensure consistent sample collection without damaging the surface or the tool \cite{mahmoudi2025evaluation}. During training, a force-sensitive resistor (FSR) embedded within a compliant swabbing tool provides ground-truth force labels, while deployment relies solely on wrist-mounted force/torque sensing and robot kinematic states, eliminating the need for tool-integrated sensors, which has practical constraints related to sterility, disposability, limited space for electronics, and frequent tool replacement \cite{mahmoudi2025data}.

To capture the history-dependent dynamics of continuous swabbing interactions and enable generalization across previously unseen surfaces and tool compliance conditions, we introduce a context-aware few-shot continual adaptation strategy that modulates a frozen recurrent backbone using surface-specific latent embeddings via feature-wise linear modulation (FiLM). This formulation enables adaptation to new interaction conditions while preserving performance on the original task, reducing estimation error under distribution shift without catastrophic forgetting.

The primary contributions of this work are:
\begin{itemize}
    \item[(i)] A context-modulated recurrent architecture for deformable contact force estimation;
    \item[(ii)] A parameter-isolated adaptation strategy enabling few-shot cross-surface generalization; and
    \item[(iii)] Experimental validation on robotic environmental swabbing demonstrating robust force estimation without tool-integrated sensing.
\end{itemize}

\section{Related Work}

While DOM is well-surveyed \cite{zhu2022challenges, arriola2020modeling}, manipulating rigid environments through a deformable tool introduces distinct dynamics that decouple wrist-level measurements from tip-level interactions. Analytical approaches, such as model-based impedance control \cite{anand2023model}, typically rely on linear constitutive assumptions (e.g., Kelvin-Voigt models) that become intractable when tools exhibit variable stiffness or viscoelastic hysteresis.

To bypass explicit modeling, direct measurement via tool-integrated sensors at the contact interface presents a potential solution \cite{mahmoudi2025data}; however, such instrumentation is frequently precluded by practical constraints, including sensor drift, safety hazards, and limited accessibility. As a viable alternative, data-driven estimators leveraging proprioceptive and tactile signals have been proposed. Recent works have utilized multimodal sensor fusion to estimate contact forces \cite{lee2019making}, while others have demonstrated that recurrent architectures (RNNs) can effectively capture non-linear deformation signatures \cite{sundaram2019learning}. However, these supervised models implicitly assume a stationary physical distribution. When subjected to domain shift such as transitioning from elastic wood to viscoelastic plastic, standard deep learning models suffer severe performance degradation, necessitating retraining that is often data-inefficient and risks catastrophic forgetting \cite{kirkpatrick2017overcoming}.

\begin{figure*}[t]
    \centering
    \includegraphics[width=\textwidth]{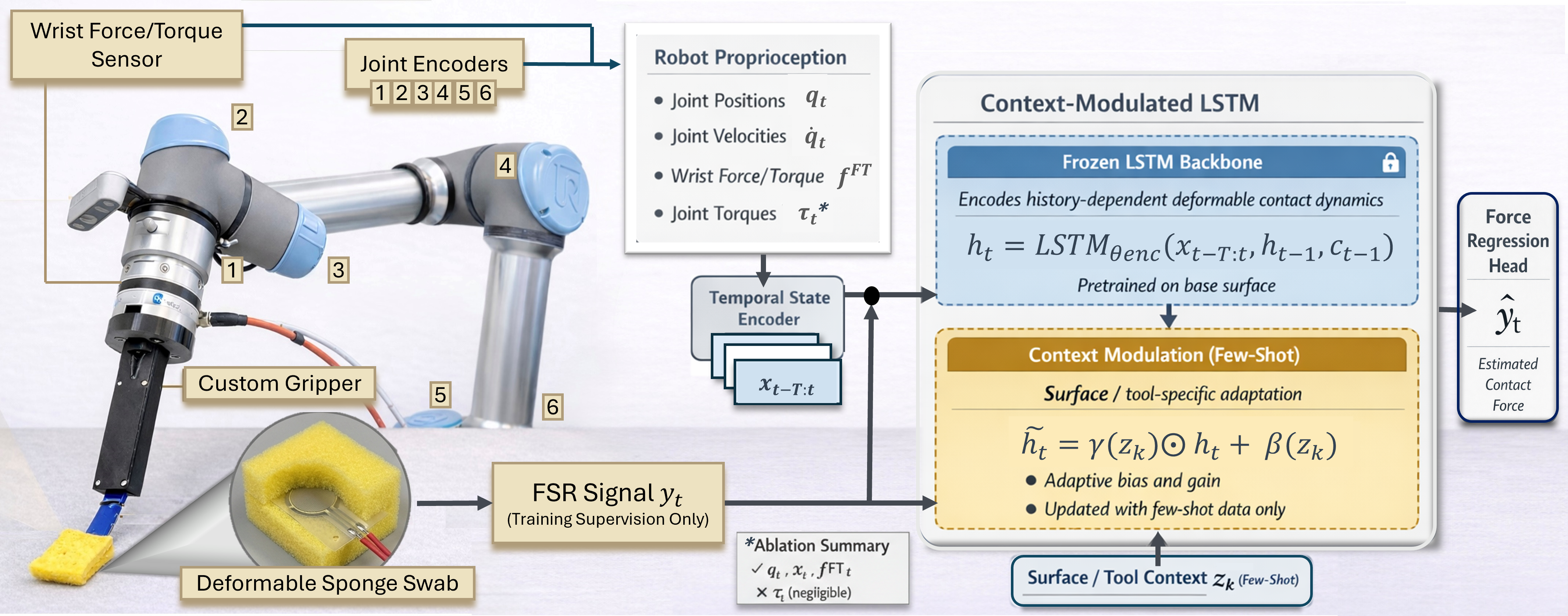}
    \caption{Overview of the proposed force estimation pipeline: robot proprioceptive states $\mathbf{x}_t$ are encoded over a temporal window $\mathbf{x}_{t-T:t}$ and processed by a context-modulated LSTM with a frozen backbone and few-shot context $\mathbf{z}_k$ to estimate contact force $\hat{y}_t$.}
    \label{fig:overview}
    \vspace{-15pt}
\end{figure*}
To address non-stationary environments, robotics research has adopted meta-learning (e.g., MAML \cite{finn2017model}) and online system identification \cite{zhu2023few} for rapid dynamics adaptation. However, standard fine-tuning often overwrites prior knowledge \cite{parisi2019continual}. Recent work has also demonstrated the effectiveness of data-driven model in updating motion model during deformable linear objects manipulation \cite{choi2026time}. While effective for online adaptation, these methods focus on single-task model updating and do not explicitly consider mechanisms for preserving performance across previously encountered interaction conditions. 

To mitigate this issue, advances in parameter--efficient transfer learning, most notably Visual Prompt Tuning (VPT) \cite{jia2022visual} and Learning to Prompt (L2P) \cite{wang2022learning}, propose freezing a pretrained backbone network and adapting to new tasks by optimizing only a small set of learnable context parameters. These approaches have shown strong performance in continual learning scenarios in vision and language domains \cite{liu2017global}. However, such prompt-based or context-isolated adaptation strategies remain largely unexplored in robotic force estimation and DTM. Our work bridges this gap by proposing a context-aware LSTM framework that treats surface and tool interaction properties as learnable latent prompts, enabling few-shot adaptation to viscoelastic environments while strictly preserving source-domain performance.

\section{Problem Formulation \& System Overview}
\label{sec:III}
We formulate contact force estimation in deformable swab tool manipulation as a supervised sequence-to-scalar regression problem, aiming to predict tip-level forces from proprioceptive measurements without relying on explicit constitutive models or permanent sensors. As illustrated in Fig. \ref{fig:overview}, the experimental system comprises a 6-DOF UR5e manipulator equipped with a custom gripper holding a disposable cellulose sponge swab (3M Corp., Maplewood, MN, USA). To provide ground-truth supervisory signals during training, a single-axis force-sensitive resistor (ThruMode FSR; Sensitronics, WA, USA) is embedded within the sponge body. This FSR was calibrated under standardized moisture conditions (1 mL deionized water, mimicking environmental sampling protocols \cite{abc2024,mahmoudi2025evaluation}) using a NIST-traceable TMS–Pro Advanced Texture Analyzer (Food Technology Corporation, Sterling, VA, USA), where a 5th-order polynomial mapping achieved a calibration RMSE $<$ 0.08 N over the operational range 0–5 N.

Training data is collected using a controlled swabbing trajectory with predefined motion parameters designed to excite viscoelastic and hysteretic deformation modes that arise during realistic surface sampling. The trajectory induces coupled normal compression, shear, and bending while maintaining approximate tip-level contact, generating force responses that depend on both instantaneous configuration and deformation history. This design yields rich, non-Markovian interaction dynamics representative of real swabbing tasks.

At each timestep $t$, the robot provides a 24-dimensional proprioceptive measurement vector: $\mathbf{x}_t
=
\begin{bmatrix}
\mathbf{q}_t &
\dot{\mathbf{q}}_t &
\boldsymbol{\tau}_t &
\mathbf{f}^{\mathrm{FT}}_t
\end{bmatrix}
\in \mathbb{R}^{24},$ where $\mathbf{q}_t, \dot{\mathbf{q}}_t \in \mathbb{R}^6$ denote joint positions and joint velocities, respectively,
$\boldsymbol{\tau}_t \in \mathbb{R}^6$ are the commanded joint torques, and $
\mathbf{f}^{\mathrm{FT}}_t
=
\begin{bmatrix}
\mathbf{F}^{\mathrm{FT}}_t &
\boldsymbol{\tau}^{\mathrm{FT}}_t
\end{bmatrix}
\in \mathbb{R}^6 $ represents the wrist-mounted force--torque sensor measurements, comprising forces
$\mathbf{F}^{\mathrm{FT}}_t \in \mathbb{R}^3$ and torques
$\boldsymbol{\tau}^{\mathrm{FT}}_t \in \mathbb{R}^3$
(OnRobot HEX-E, force resolution $0.5\,\mathrm{N}$ and torque resolution $0.025\,\mathrm{Nm}$).
All signals are sampled synchronously at 300 Hz and time-aligned with the FSR supervision signal.

Due to viscoelastic hysteresis and stress relaxation, the contact force at time $t$ cannot be inferred from instantaneous measurements alone. Instead, contact force is modeled as a nonlinear function of temporal interaction history: $
y_t = f_{\theta}\!\left(\mathbf{x}_{t-T:t}, \mathbf{z}_k\right),
\label{eq:force_estimation}
$ where $\mathbf{x}_{t-T:t} \in \mathbb{R}^{T \times 24}$  denotes a sliding temporal window of robot states, $\theta$ represents
the learnable model parameters, and $\mathbf{z}_k \in \mathbb{R}^{d_z}$ is a task-specific latent context embedding encoding effective material interaction properties of surface or tool $k$ (e.g., stiffness, friction, and damping). For the baseline surface, the context vector is learned jointly with $\theta$ during pretraining. For unseen surfaces
($k \neq 0$), the corresponding context vectors $\mathbf{z}_k$ are learned
via few-shot adaptation while keeping $\theta$ frozen, enabling
cross-domain generalization without catastrophic forgetting.

Accordingly, the estimator $f_\theta$ satisfy three requirements: (i) capture long-range temporal dependencies (up to $2\,\mathrm{s}$) to model stress relaxation; (ii) support real-time inference compatible with high-frequency control; and (iii) enable parameter-isolated adaptation across interaction conditions.

Figure~\ref{fig:overview} summarizes the complete data flow. During training, synchronized proprioceptive and FSR measurements are segmented into overlapping temporal windows, with the FSR providing supervisory force labels. At deployment, the trained estimator $f_\theta(\cdot, \mathbf{z}_k)$ predicts contact forces using only robot state measurements, enabling sterile operation with disposable tools and no embedded sensing. 

\section{CONTEXT-MODULATED RECURRENT ESTIMATION FRAMEWORK}

To address the stochasticity and non-linear transmission dynamics inherent in a deformable tool, a two-stage learning framework is proposed. First, a recurrent neural architecture is identified to model the history-dependent viscoelasticity of the tool–surface interface. Second, a parameter-isolated adaptation strategy is introduced to resolve domain shifts caused by material variations, ensuring robust generalization without catastrophic forgetting.

\subsection{Recurrent Modeling of Contact Dynamics}
Building upon the problem formulation in Sec. \ref{sec:III}, the objective is to approximate the mapping function $f_\theta$ governing the temporal evolution of contact stress induced by internal tool deformation. Because the viscoelastic hysteresis, stress relaxation, and rate-dependent stiffness exhibited in the deformable tools \cite{choi2026time}, the force at time $t$ depends not only on the current configuration, but also on the accumulated deformation history. 

Following a comparative evaluation of temporal modeling strategies (Section \ref{sec:result}), a recurrent model was selected to capture the non-Markovian dynamics inherent to viscoelastic contact \cite{choi2026time}. The observed path-dependent energy dissipation and phase-lag effects require a modeling framework capable of maintaining an evolving internal state representing deformation history \cite{lara2020temporal}. Accordingly, a Long Short-Term Memory (LSTM) network \cite{graves2012long} is adopted as the contact dynamics encoder due to its ability to maintain long-range temporal dependencies through gated memory mechanisms. Unlike standard recurrent units, the LSTM utilizes explicit gating mechanisms \cite{gers2000learning} to regulate the retention and forgetting of past information. This structure enables the network to selectively preserve deformation-induced stress components while discarding transient noise, closely mirroring the physical accumulation and relaxation processes characteristic of viscoelastic contact mechanics \cite{qin2024physics, benabou2021development}.

Let the observable robot state at time $t$ be denoted by $\mathbf{x}_t \in \mathbb{R}^{24}$. Given a temporal input sequence $\mathbf{x}_{t-T:t} = \{\mathbf{x}_{t-T}, \ldots, \mathbf{x}_t\}$ of length $T$, the LSTM backbone computes a hidden state $\mathbf{h}_t$ and cell state $\mathbf{c}_t$ according to: 
\begin{equation}
    \mathbf{h}_t, \mathbf{c}_t = \mathrm{LSTM}_{\theta_{\mathrm{enc}}}(\mathbf{x}_t, \mathbf{h}_{t-1}, \mathbf{c}_{t-1}),
\end{equation}
where $\theta_{\mathrm{enc}}$ denotes the parameters of the recurrent backbone. The hidden state $\mathbf{h}_t$ represents a compact latent encoding of accumulated deformation energy and delayed stress responses over the input sequence.
Let $y_t \in \mathbb{R}$ denote the ground-truth tip-level contact force obtained from the embedded tactile sensor during training. The predicted force $\hat{y}_t$ is obtained by passing the latent representation through a regression head,
\begin{equation}
\hat{y}_t = g_{\phi}(\mathbf{h}_t),
\end{equation}
where $g_{\phi}$ is a parameterized mapping from the latent state to a scalar force estimate. This sequence-to-scalar formulation captures the history-dependent nature of viscoelastic contact without relying on instantaneous state–force relationships.

The recurrent estimator is implemented as a two-layer LSTM architecture with 64 hidden units per layer. To map the high-dimensional temporal features to a physical force estimate, the final hidden state is processed by a two-layer feedforward regression head consisting of a 32-unit intermediate layer and a Rectified Linear Unit (ReLU) activation. The complete parameter set is $\Theta = \{\theta_{\mathrm{enc}}, \phi\}$. Model training on the source domain minimizes the mean squared error between predicted and ground-truth contact forces. Owing to its compact parameterization, the resulting estimator achieves sub-millisecond inference latency, enabling real-time deployment within high-frequency estimation and feedback control loops.

\subsection{Parameter-Isolated Few-Shot Adaptation via Context Modulation}

While the recurrent backbone captures the generic history-dependent deformation dynamics of the swabbing tool, the constitutive properties of the tool--surface interaction, including effective stiffness, damping, and frictional dissipation, vary substantially across contact conditions. These variations induce a systematic domain shift that cannot be captured by a single static parameterization without retraining or sacrificing performance on previously encountered interactions. To address this challenge, we introduce a parameter-isolated few-shot adaptation mechanism that modulates latent dynamics through low-dimensional conditioning while preserving shared temporal representations.

Let $\mathbf{h}_t \in \mathbb{R}^{d_h}$ denote the hidden state produced by the frozen LSTM encoder after processing the temporal window $\mathbf{x}_{t-T:t}$, where $d_h = 128$. Adaptation to a new interaction condition $k$ is achieved via a low-dimensional latent context vector $\mathbf{z}_k \in \mathbb{R}^{d_z}$, with $d_z \ll d_h$. Rather than retraining the encoder, the context vector modulates the latent representation using FiLM \cite{perez2018film}, yielding a context-conditioned state:
\begin{equation}
    \tilde{\mathbf{h}}_t = \gamma(\mathbf{z}_k) \odot \mathbf{h}_t + \beta(\mathbf{z}_k),
    \label{eq:film_modulation}
\end{equation}
where $\odot$ denotes element-wise multiplication. This conditioning mechanism enables surface-specific modulation of the latent deformation representation without altering recurrent dynamics.
The modulation functions $\gamma(\cdot)$ and $\beta(\cdot)$ are implemented as lightweight MLP projectors that map $\mathbf{z}_k$ to scale and bias vectors in $\mathbb{R}^{d_h}$. Each projector consists of a two-layer network ($d_z \rightarrow 64 \rightarrow d_h$) with ReLU activations. FiLM conditioning thus provides a low-rank modulation mechanism that enables expressive domain adaptation without increasing model depth or modifying shared parameters. This conditioning mechanism is conceptually related to task-conditioned parameterization approaches explored in continual learning literature, where task embeddings are used to modulate network behavior without retraining shared representations \cite{von2019continual}. The conditioned latent state $\tilde{\mathbf{h}}_t$ is then passed to the frozen regression head $g_\phi(\cdot)$ to produce the force estimate $\hat{y}_t = g_\phi(\tilde{\mathbf{h}}_t).$

Following pretraining on the source domain, the recurrent encoder parameters $\theta_{\mathrm{enc}}$, FiLM projector parameters, and regression head parameters $\phi$ are permanently frozen. Adaptation to a novel interaction condition is performed using a small support set $\mathcal{S}_k$ consisting of $K$ swabbing trials, during which only the latent context vector $\mathbf{z}_k$ is optimized:
\begin{equation}
\begin{aligned}
    \mathbf{z}_k^* &= \arg\min_{\mathbf{z}_k} \frac{1}{|\mathcal{S}_k|} \sum_{(\mathbf{x}_{t-T:t}, y_t)\in \mathcal{S}_k} \\
    &\quad \Big\| y_t - g_\phi \big( \gamma(\mathbf{z}_k)\odot \mathrm{LSTM}_{\theta_{\mathrm{enc}}}(\mathbf{x}_{t-T:t}) + \beta(\mathbf{z}_k) \big) \Big\|^2.
\end{aligned}
\end{equation}
The context vector $\mathbf{z}_k$ is initialized to zero at adaptation time. To ensure stable behavior prior to adaptation, the final layer of the $\gamma$ projector is initialized with a bias of $+1$, such that the initial modulation corresponds to the identity transform ($\gamma \approx \mathbf{1}$, $\beta \approx \mathbf{0}$).

Each interaction condition is associated with a distinct latent embedding $\mathbf{z}_k$, and learned context vectors are stored in a domain-indexed context bank $\mathcal{Z} = \{\mathbf{z}_0, \mathbf{z}_1, \ldots, \mathbf{z}_K\}$. During inference, adaptation corresponds to selecting the appropriate context embedding without modifying shared network parameters. Because all shared parameters remain fixed after pretraining, previously learned domain mappings remain invariant under subsequent adaptations, thereby eliminating interference between domains. This separation between shared dynamics and domain-specific conditioning provides an explicit parameter isolation mechanism that prevents interference between tasks. This formulation casts adaptation as conditioning in latent space rather than parameter re-estimation, enabling rapid few-shot calibration while maintaining constant model capacity. Consequently, the framework supports continual adaptation across material domains with negligible memory overhead proportional only to the dimensionality of the context vectors.

\section{EXPERIMENTAL EVALUATION}
\label{sec:result}
This section evaluates the proposed context-modulated recurrent estimation framework through a series of controlled experiments designed to assess modeling accuracy, data efficiency, sensing sensitivity, and robustness to domain shifts. The evaluation focuses on four key aspects: (i) the suitability of recurrent architectures for modeling history-dependent force transmission in deformable tools, (ii) data efficiency during training on a baseline surface, (iii) sensitivity to different proprioceptive sensing modalities, and (iv) robustness to material-induced domain shifts through parameter-isolated few-shot adaptation. All experiments are conducted using identical motion trajectories and synchronized sensor measurements to ensure a fair comparison between models and conditions.

\subsection{Comparative Analysis of Temporal Architectures}
 Table~\ref{tab:architecture_comparison}  summarizes the predictive performance of the five evaluated architectures. To ensure a rigorous evaluation, all models were trained using identical optimization settings with strict trial-level splitting, guaranteeing that the validation metrics reflect true generalization to unseen physical interactions rather than memorization of overlapping windows. Reported metrics represent the mean performance over five random initialization seeds. The standard deviation of RMSE across five seeds remained below 0.8 ADC for all architectures, indicating stable convergence.

\begin{table}[h]
    \centering
    \begin{threeparttable}
        \caption{Comparison of temporal modeling architectures for contact force estimation on the baseline wood surface.}
        \label{tab:architecture_comparison}
        \setlength{\tabcolsep}{4pt}
        \renewcommand{\arraystretch}{1.15}
        
        \begin{tabularx}{\linewidth}{l >{\centering\arraybackslash}X >{\centering\arraybackslash}X >{\centering\arraybackslash}X >{\centering\arraybackslash}X}
            \hline
            \textbf{Model} & \textbf{Params} & \textbf{Latency (ms)} & \textbf{RMSE} $\downarrow$ & $\mathbf{R^2}$ $\uparrow$ \\
            \hline
            MLP (Baseline) & 746,497 & \textbf{0.086} & 90.28 & 0.923 \\
            CNN (1D-Conv) \cite{ferretti2020towards} & 69,633 & 0.418 & 78.23 & 0.942 \\
            TCN \cite{BaiTCN2018}& 133,953 & 1.613 & 35.31 & 0.988 \\
            Transformer \cite{10.1145/3447548.3467401}& 104,129 & 0.754 & 30.82 & 0.991 \\
            \textbf{LSTM (Ours)} & \textbf{60,545} & 0.371 & \textbf{26.99} & \textbf{0.992} \\
            \hline
        \end{tabularx}
        
        \begin{tablenotes}[flushleft]
            \footnotesize
            \item \textit{All models were trained and evaluated on an Intel Core i9-13900H CPU. Latency is reported under single-threaded execution to reflect real-time deployment conditions. RMSE and $R^2$ calculated on a fixed validation set of 14 unseen trials.}
        \end{tablenotes}
    \end{threeparttable}
    
\end{table}

The quantitative results confirm the efficacy of the recurrent approach. The LSTM achieves the lowest estimation error (RMSE = 26.99) and highest coefficient of determination ($R^2 = 0.992$), outperforming the Transformer by approximately 12.4\% and the TCN by over 23.5\%. While the Transformer effectively captures global temporal dependencies, it does not provide a measurable accuracy gain relative to LSTM in this regime, despite higher computational overhead. A Wilcoxon signed-rank test confirmed that this performance improvement over the Transformer is statistically significant ($p = 0.049, \alpha = 0.05$) across the 14 validation trials.
In contrast, the LSTM’s recurrent cell state ($c_t$) functions as an efficient memory accumulator, aligning with the continuous energy integration inherent to viscoelastic dynamics. Crucially, the LSTM achieves this accuracy with sub-millisecond inference latency (0.37 ms) and a compact parameter footprint (60k), providing the optimal balance for real-time control loops where strict timing guarantees are required.

\begin{figure}[h]
    \centering
    \includegraphics[width=1\linewidth]{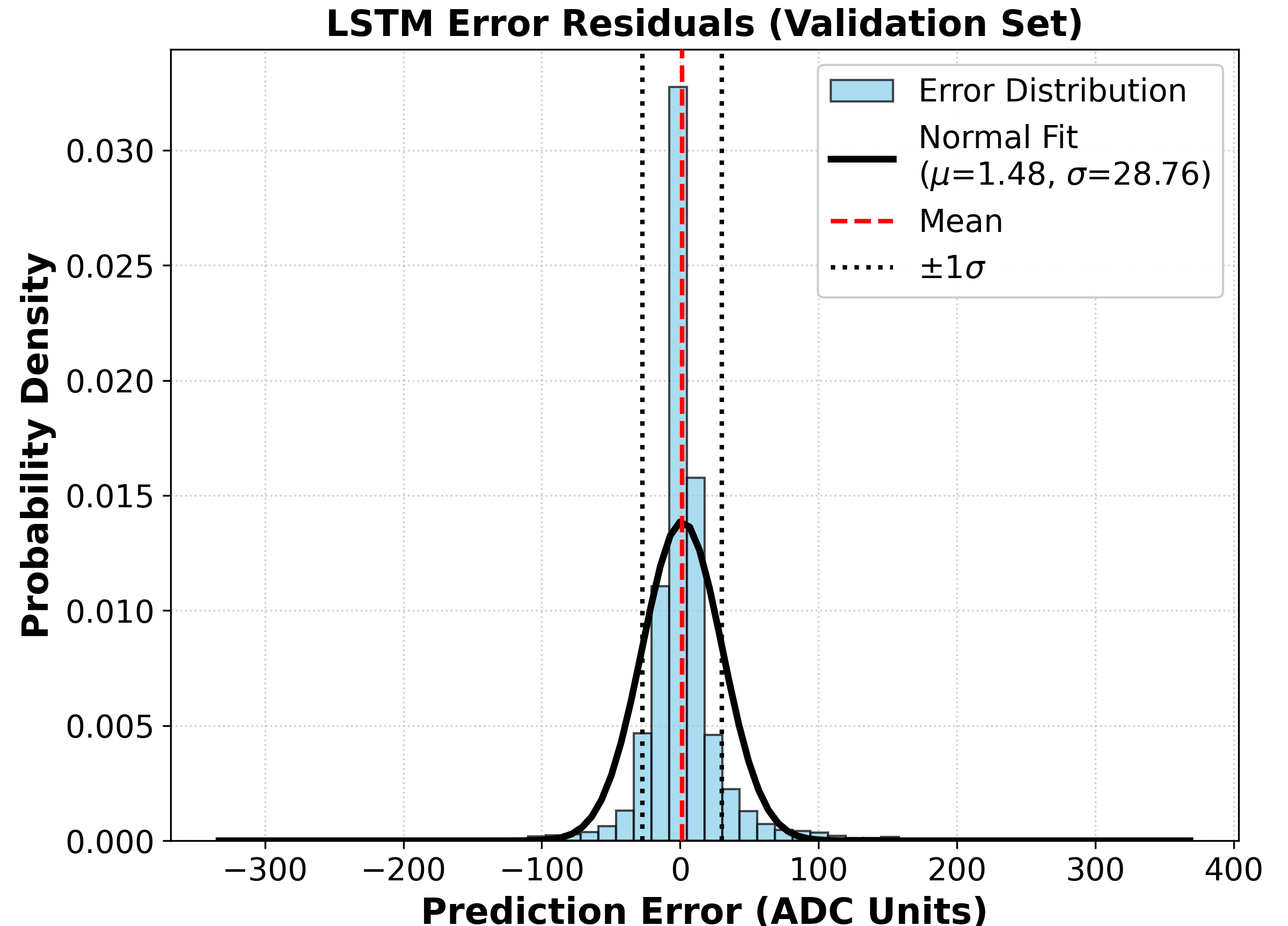}
    \vspace{-10pt}
    \caption{LSTM error residual distribution on the validation set ($N=14$). }
    \label{fig:error_hist}
    \vspace{-5pt}
\end{figure}

Fig.~\ref{fig:error_hist} further analyzes the error residuals. The distribution is symmetric and centered near zero ($\mu = 1.48$ ADC units), confirming that the LSTM provides unbiased estimates without systematic drift. The distribution is sharply peaked around zero, indicating that most prediction errors are small in magnitude.
\vspace{-10pt}

\begin{figure}[h]
    \centering
    \includegraphics[width=1\linewidth]{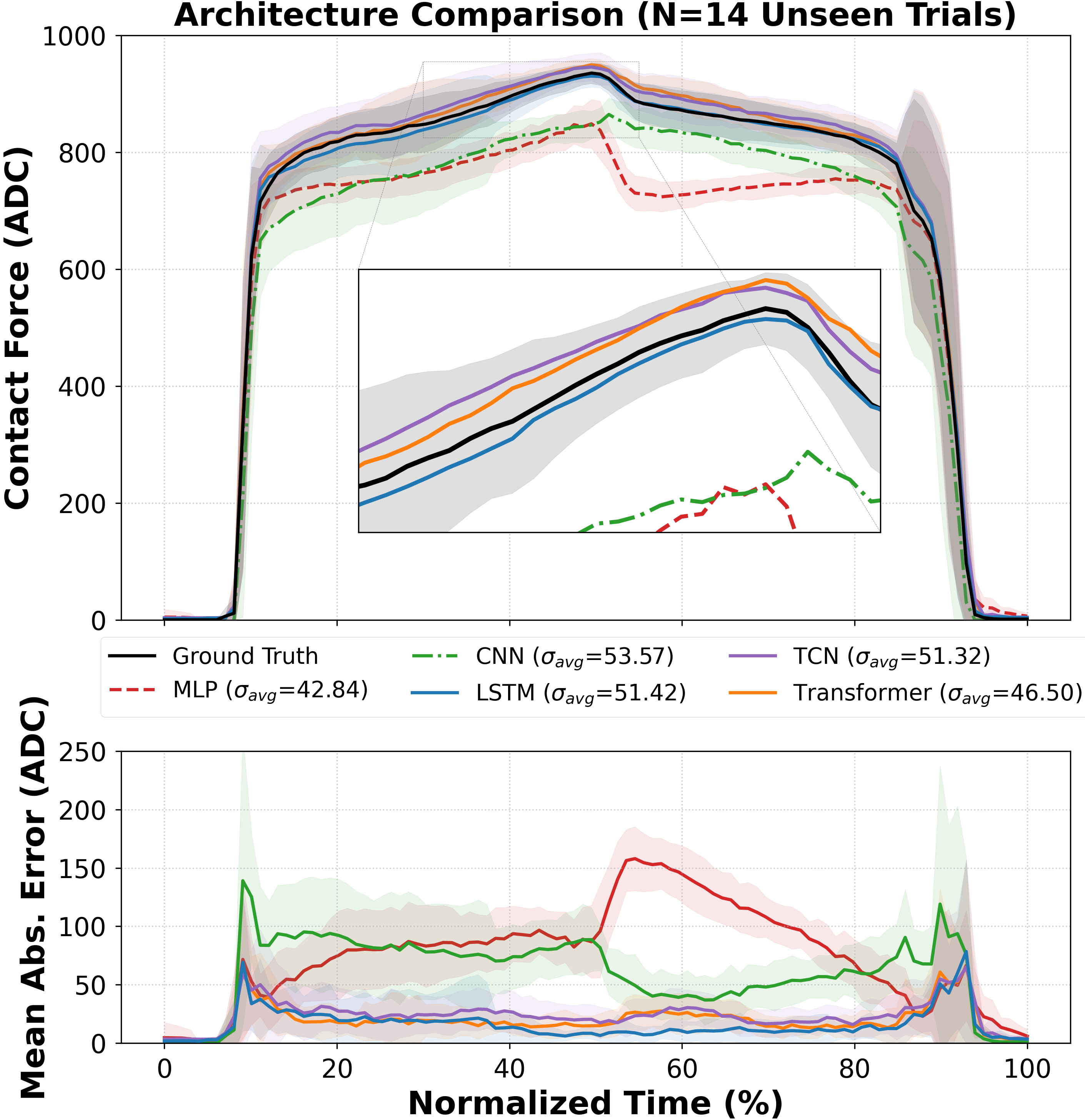}
    \caption{Architecture comparison on 14 unseen trials. Top: mean predicted contact force profiles with confidence bands. Bottom: mean absolute error across normalized time.}
    \label{fig:architecture_comparison}
\end{figure}

Fig.~\ref{fig:architecture_comparison} visualizes this performance on the validation set. Solid lines represent the mean force profile across 14 unseen trials, while shaded regions denote $\pm 1$ standard deviation, illustrating the model's consistency. The LSTM tracks the ground truth closely during both the sharp impact onset and the non-linear relaxation phase. The lower panel confirms that the LSTM maintains the lowest mean absolute error in the critical 50--70\% transition region, where temporal memory is most active. Consequently, the LSTM is selected as the backbone for the subsequent robustness experiments. 

\begin{figure*}[t]
    \centering
    \includegraphics[width=1\linewidth]{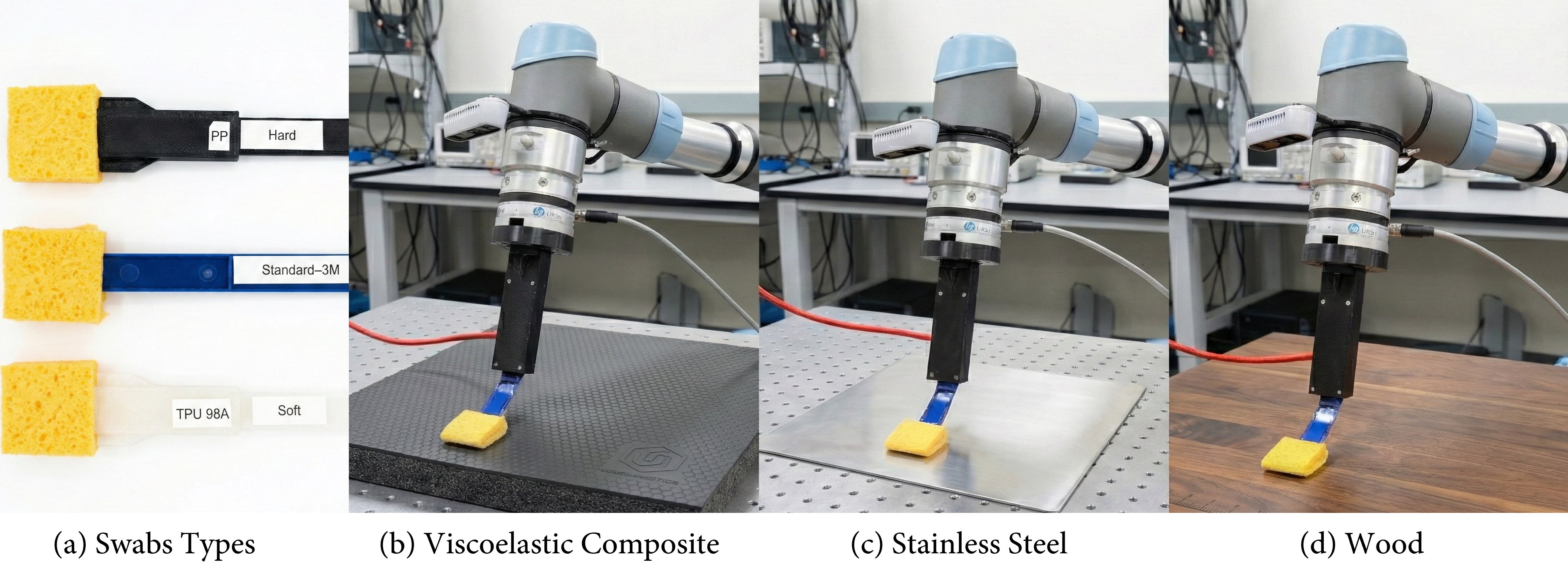}
    \caption{Experimental conditions used to evaluate surface- and tool-induced domain shifts. 
    (a) Viscoelastic composite surface exhibiting high damping and hysteresis. 
    (b) Stainless steel surface representing near-rigid contact. 
    (c) Wood surface used as the baseline training condition. 
    (Bottom) Three swab tools with distinct compliance levels: rigid polypropylene (Hard), standard cellulose sponge (3M), and compliant TPU-98A (Soft). 
    All experiments were performed using identical motion trajectories to isolate constitutive differences in tool--surface interaction dynamics.}
    \label{fig:experimental_conditions}
\end{figure*}
\subsection{Data Efficiency and Learning Dynamics}
To evaluate sample efficiency, the LSTM model was trained on progressively increasing fractions of the baseline wood dataset, ranging from 10\% (5 trials) to 100\% (56 trials). All experiments employed identical hyperparameters and strict trial-level splits to prevent temporal leakage. Table \ref{tab:data_efficiency} details the impact of training volume on prediction accuracy.
\begin{table}[h]
\centering
\caption{Impact of Training Data Volume on Prediction Accuracy}
\label{tab:data_efficiency}
\setlength{\tabcolsep}{6pt}
\renewcommand{\arraystretch}{1.1}
\begin{tabular}{ccccc}
\toprule
\textbf{Data Fraction} & \textbf{Trials} & \textbf{RMSE} $\downarrow$ & \textbf{MAE} $\downarrow$ & \textbf{$\mathbf{R^2}$} $\uparrow$ \\
\midrule
10\% & 5 & 43.22 & 28.71 & 0.9763  \\
15\% & 8 & 50.23 & 33.95 & 0.9702 \\
30\% & 16 & 41.93 & 25.35 & 0.9804 \\
50\% & 28 & 35.53 & 21.21 & 0.9821 \\
\textbf{100\%} & \textbf{56} & \textbf{26.99} & \textbf{14.64} & \textbf{0.9922} \\
\bottomrule
\end{tabular}
\vspace{-15pt}
\end{table}

The model achieves a high $R^2$ (0.976) with only five training trials (10\% of data). This suggests that dominant quasi-static force components are rapidly captured from limited exposure.
However, RMSE reveals a more complex learning process. The error initially increases to the 15\% mark (RMSE 43.22 $\to$ 50.23), indicating that the initial five trials did not capture the full variance of the manual swabbing distribution. As data volume exceeds 30\%, the model begins to effectively generalize to these variations, with full-data training ultimately reducing the RMSE by approximately 38\% relative to the 10\% training condition. This confirms that, while the LSTM captures the dominant elastic response rapidly, a broader dataset is required to resolve the subtler, history-dependent viscoelastic effects.

\subsection{Sensitivity to Proprioceptive Modalities}
Having established the data volume required for robust generalization, we next investigate the contribution of individual proprioceptive modalities to the estimation task. While the full state vector provides a comprehensive description of the robot's dynamics, it is critical to determine which specific features, kinematics versus kinetics, drive the predictions and whether certain inputs introduce detrimental noise. Ablation experiments were conducted by selectively removing input channels from the 24-dimensional state vector.

\begin{table}[h]
\begin{threeparttable}
\centering
\caption{Ablation Study of Input Modalities}
\label{tab:ablation}
\setlength{\tabcolsep}{16pt} 
\renewcommand{\arraystretch}{1.15}
\begin{tabular}{lcc}
\hline
\textbf{Configuration} & \textbf{Input Dim} & \textbf{RMSE} $\downarrow$ \\
\hline
Baseline (Full State) & 24 & 26.99 \\
No Velocity & 18 & 30.40 \\
\textbf{No Joint Torque} & \textbf{18} & \textbf{26.39} \\
No F/T Sensor & 18 & 65.63 \\
\hline
\end{tabular}

        \begin{tablenotes}[flushleft]
            \footnotesize
            \item \textit{Note: Removing noisy joint torque estimates slightly improves accuracy compared to the baseline, while removing the F/T sensor more than doubles the error.}
        \end{tablenotes}
\end{threeparttable}
\vspace{-15pt}
\end{table}

Removing wrist force–torque measurements increases RMSE by more than 143\%, confirming that deformation-induced wrench signals are essential for force reconstruction. Excluding joint velocities produces a moderate degradation in performance, indicating that velocity contributes to modeling rate-dependent stiffness and damping. Interestingly, removing commanded joint torques yields marginal improvement, suggesting that torque signals introduce noise or redundant information in this setting.
These findings indicate that accurate force estimation depends primarily on kinematics and external wrench measurements, while internal torque commands provide limited additional value.

\subsection{Robustness to Material- and Tool-Induced Domain Shift}

\label{sec:robustness}

To assess generalization limits, robustness was evaluated under  variations in environmental constitutive properties and tool compliance. This experiment determines whether the learned recurrent latent representation encodes transferable contact mechanics or is over-specialized to the baseline configuration. Experimental conditions are shown in Fig.~\ref{fig:experimental_conditions}.

\subsubsection{Experimental Setup}
Three swab tools with identical geometry and sponge dimensions were fabricated to isolate bending stiffness $k_{tool}$ as the sole structural variable: (a)\textbf{Standard–3M}: Commercial baseline tool (training domain). (b) \textbf{Hard (PP)}: 3D-printed polypropylene stem with increased stiffness. (c) \textbf{Soft (TPU 98A)}: 3D-printed thermoplastic polyurethane stem with reduced elastic modulus. Each tool was evaluated on three surfaces spanning distinct constitutive regimes: (a) \textbf{Wood}: Moderately elastic (training surface). (b) \textbf{Stainless Steel}: Near-rigid interface ($k_{env} \gg k_{sponge}$). (c) \textbf{Viscoelastic Composite}: Foam-backed synthetic surface with elevated damping ($c_{env}$) and hysteresis. These yield a $3 \times 3$ interaction matrix (nine regimes). The LSTM backbone was trained exclusively on Standard–3M interacting with Wood and evaluated under (i) zero-shot transfer and (ii) parameter-isolated few-shot adaptation (5 trials updating only context vector and readout). Results are summarized in Table~\ref{tab:comprehensive_robustness}.

\begin{table*}[t]
\centering
\caption{
Evaluation of surface- and tool-induced domain shift 
(mean $\pm$ std over 5 random seeds). RMSE is reported in ADC units; 
1 N $\approx$ 120 ADC within the calibrated operating region. 
$R^2$ corresponds to the few-shot adapted model unless otherwise noted.
}
\label{tab:comprehensive_robustness}
\setlength{\tabcolsep}{6pt}
\renewcommand{\arraystretch}{1.15}
\begin{tabular}{ll c c c c}
\hline
\textbf{Surface Domain} & \textbf{Swab Tool} 
& \textbf{Zero-Shot RMSE} 
& \textbf{Few-Shot RMSE} 
& \textbf{$R^2$ (Few-Shot)} 
& \textbf{Improvement} \\
 &  & (ADC $\downarrow$) & (ADC $\downarrow$) & ($\uparrow$) & (\%) \\
\hline
\hline

\multirow{4}{*}{\textbf{Wood (Baseline)}} 
& Standard (Train Domain) 
& $26.9 \pm 1.8$ 
& -- 
& $0.992$ 
& -- \\

& Hard (PP) 
& $44.1 \pm 2.6$ 
& $29.4 \pm 2.1$ 
& $0.989$ 
& $33.3\%$ \\

& Soft (TPU) 
& $39.5 \pm 2.3$ 
& $30.2 \pm 2.0$ 
& $0.988$ 
& $23.5\%$ \\

& \textbf{Wood (Retained After All Adaptation)} 
& -- 
& $\mathbf{27.1 \pm 1.9}$ 
& $\mathbf{0.991}$ 
& \textbf{No Forgetting} \\

\hline

\multirow{3}{*}{\textbf{Stainless Steel}} 
& Standard 
& $40.2 \pm 2.4$ 
& $30.5 \pm 2.2$ 
& $0.987$ 
& $24.1\%$ \\

& Hard (PP) 
& $58.9 \pm 3.1$ 
& $33.6 \pm 2.7$ 
& $0.982$ 
& $42.9\%$ \\

& Soft (TPU) 
& $35.6 \pm 2.0$ 
& $29.3 \pm 1.9$ 
& $0.989$ 
& $17.7\%$ \\

\hline

\multirow{3}{*}{\textbf{Viscoelastic Composite}} 
& Standard 
& $88.4 \pm 4.1$ 
& $33.8 \pm 2.8$ 
& $0.979$ 
& $61.8\%$ \\

& Hard (PP) 
& $95.7 \pm 4.8$ 
& $36.2 \pm 3.0$ 
& $0.975$ 
& $62.2\%$ \\

& Soft (TPU) 
& $65.8 \pm 3.2$ 
& $30.9 \pm 2.4$ 
& $0.984$ 
& $53.0\%$ \\

\hline
\end{tabular}
\vspace{-15pt}
\end{table*}

\subsubsection{Mechanical Interpretation}
On Stainless Steel, zero-shot RMSE increases by approximately 45\%, reflecting stiffness mismatch ($k_{env} \uparrow$). On the Viscoelastic Composite, error increases by more than 200\%, indicating that damping-induced phase lag and hysteresis cannot be captured without recalibration. Tool compliance introduces a secondary shift. On Wood, both Hard and Soft stems increase zero-shot error, confirming sensitivity to $k_{tool}$. The most severe degradation occurs for the Hard (PP) tool on Stainless Steel, where rigid–rigid interaction amplifies high-frequency impact transients. The Soft (TPU) tool consistently exhibits lower zero-shot error on rigid surfaces, suggesting that added compliance passively filters impact spikes and partially aligns the signal distribution with the training regime.

Across all nine regimes, few-shot adaptation reduces RMSE by 18–63\%, restoring performance near baseline levels. On the Viscoelastic Composite surface, error reductions exceed 60\%, demonstrating that the latent context vector compensates effectively for changes in $k_{env}$ and $c_{env}$ without modifying the recurrent backbone.
Performance on the original Wood domain remains statistically unchanged after all adaptation cycles (27.1 $\pm$ 1.9 ADC, $R^2 = 0.991$), confirming structural prevention of catastrophic forgetting.

\section{Conclusion}

This paper presented a context-modulated recurrent framework for estimating contact forces during DTM. By modeling deformation-history dynamics with a compact LSTM backbone and separating domain-specific interaction properties through parameter-isolated context embeddings, the proposed approach enables accurate force estimation without tool-integrated sensing.

Experimental evaluation across multiple surfaces and tool compliance conditions demonstrated strong robustness under domain shift, with substantial error reductions achieved through few-shot adaptation while preserving baseline performance without catastrophic forgetting. These results indicate that deformation-history representations are transferable across interaction conditions, while low-dimensional conditioning effectively captures variations in material properties. Future work will explore automatic context inference, integration with adaptive control strategies, and deployment in unstructured environments. The proposed framework provides a scalable foundation for reliable force estimation in robotic surface interaction tasks involving compliant tools.





\section*{ACKNOWLEDGMENT}
 The authors acknowledge the use of Grammarly for enhancing the grammar and flow of the manuscript's text, as well as Adobe Photoshop's AI editor for improving the background and overall image quality of Figures \ref{fig:rvsf}, \ref{fig:overview}, and \ref{fig:experimental_conditions}.


\bibliographystyle{IEEEtran}
\bibliography{IEEEtranBST/IEEEexample}

\end{document}